\documentclass[letterpaper, 10 pt, conference]{IEEEconf}      
 \IEEEoverridecommandlockouts                              
\usepackage{graphicx}
\usepackage{float}
\graphicspath{ {images/} }
\usepackage{caption}
\usepackage{subcaption}
\usepackage{amsmath}
\usepackage{cite}
\usepackage{algorithm}
\usepackage{algpseudocode}
\usepackage{url}
\usepackage{hyperref}
\usepackage{xcolor}
\usepackage{comment}
\usepackage{array}
\usepackage{amsmath,amssymb}
\usepackage{booktabs} 
\usepackage{censor}

\newcommand{\orcid}[1]{\href{https://orcid.org/#1}{\includegraphics[width=10pt]{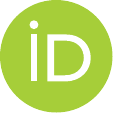}}}

\begin{document}

\title{\Large\bf Modeling Branches for Active Manipulation using Iterative Parameter Estimation \\
\author{ Madhav Rijal$^{1}$ \orcid{0000-0001-9400-0877}, Rashik Shrestha \orcid{0009-0007-7939-2623}, Trevor Smith\orcid{0000-0002-8921-9281}, Yu Gu \orcid{0000-0003-3165-3269}\vspace{2mm}\\
\textit{West Virginia University}\\
Morgantown, USA}

\thanks{The authors are with the Department of Mechanical and Aerospace Engineering, West Virginia University, Morgantown, WV 26505, USA}
\thanks{ $^{1}$ Madhav Rijal is the corresponding author
{\tt\small mr00059@mix.wvu.edu}}%
\thanks{This study was supported in part by USDA NIFA Award 2022-67021-36124 "Collaborative Research: NRI: StickBug - an Effective Co-Robot for Precision Pollination", Statler fellowship Award and the National Science Foundation Graduate Research Fellowship Award \#2136524.}  

}
\maketitle

\title{\Large\bf Modeling Branches for Active Manipulation using Iterative Parameter Estimation }
\maketitle
\begin{abstract}

This study presents a method for modeling diverse plant branches by iteratively estimating material parameters to support delicate branch manipulation. Branch manipulation is necessary in agricultural robotics for plant repositioning, stabilizing, and clearing visual obstructions in dense foliage. The proposed method builds a tetrahedral branch model from point-cloud data and simulates its behavior using the finite element method. Using real observed deformation data, it iteratively estimates branch parameters and then computes an optimal path with a deformation-aware motion planner to move and stabilize branches within another robot’s field of view. Across 30 trials on branches with varying geometries and material properties, the proposed method reduced the deformation energy by 35.69$\%$ while increasing the path length by 8.10$\%$ on average.
 
 
\end{abstract}

\section{Introduction}

Manipulating branches can facilitate various agricultural tasks, including pruning, harvesting, staking, and pollinating . During these tasks, humans often move branches to stabilize the movement of agricultural products, such as fruits and flowers, for manipulation or to find hidden products within dense canopies. Additionally, the ability to manipulate branches enables navigation within agricultural environments with dense overhanging branches and vines.

Many existing agricultural robotic systems often treat branches as obstacles to be avoided \cite{yan2024obstacle}. While some existing work has employed branch manipulation, its primary objective is to move branches out of the robot’s preplanned path \cite{zhang2023push}, without accounting for branch safety. In contrast, active branch manipulation intentionally moves branches to directly facilitate the primary task, for example, revealing hidden products or moving products closer to another manipulator, while minimizing damage \cite{rijal2025force}. 
Existing approaches to active manipulation arbitrarily manipulate branches without accounting for the mechanical stresses and deformations imposed on these delicate plant structures.

Manipulating branches using a naive heuristic, without accounting for their deformable properties, can break shoots, or damage buds and flowers, reducing the plant’s long-term productivity for growers. Most work in deformable object manipulation requires an accurate model to simulate the object's behavior for planning. However, natural branches vary widely in geometry and material properties, it is difficult to model them accurately using conventional 3D CAD tools. Therefore, manipulating living material, with spatially varying deformable properties shaped by complex geometry, requires a different approach than manipulating more uniform inanimate objects such as ropes and clothes.
 
\begin{figure}[t]
    \centering
    \includegraphics[trim=0.5cm 1cm 9cm 5cm,clip,width=0.48\textwidth]{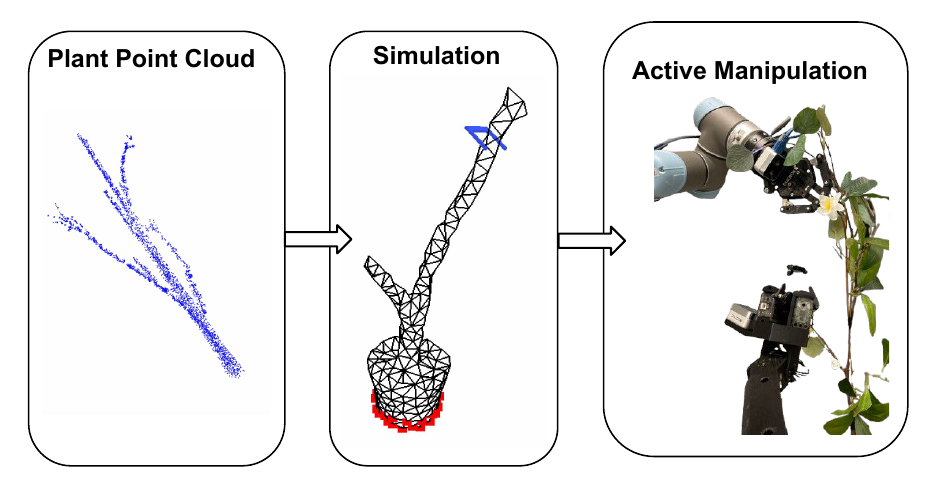}
    \caption{Overview of the proposed pipeline for active branch manipulation. An initial plant point cloud is reconstructed from perception and converted into a simulation-ready tetrahedral volumetric mesh. This deformable model is then used to plan manipulation trajectories that reposition the target such as flowers, into another robot’s reachable workspace or field of view.}
    \label{fig:intro}
\end{figure}

In addition, predicting the deformable behavior of branches remains challenging because these living materials can undergo large, highly nonlinear deformations during manipulation.\cite{zhu2022challenges}. Unlike many engineered structures that operate within small-strain regimes, branches can undergo substantial bending and geometric changes, violating linear-elastic assumptions. These violations often result in non-physical artifacts such as fictitious strain-energy accumulation and volume inconsistencies \cite{zhuang2000globaldeformation}. Additionally, approaches with position-based dynamics may exhibit numerical instability at fine spatial resolutions without intractably small time steps \cite{bender2017pbd_survey}, limiting robustness across highly variable scales of branch geometry. Alternatively, static mass-spring system models are computationally faster at capturing deformable behavior but often require extensive parameter tuning \cite{arriola2020modeling}. These issues become more pronounced in active manipulation, where contact locations and external loads vary dynamically during robot–plant interactions and cannot be specified a priori. Current limitations in branch modeling underscore the need of an efficient and physically realistic pipeline as shown in Fig. \ref{fig:intro} for large deformations under interactive forces. 

In this work, we present a novel interactive plant-simulation framework for estimating the large deformable behavior of branches. This framework integrates an automated 3D plant skeletonisation reconstructed from real-world perception data and a finite-element-method (FEM) based simulation to model the plant. In addition, a custom motion planner is developed to generate trajectories that safely manipulate branches by limiting excessive deformation and induced forces. Overall, this work's contributions include:

\begin{enumerate}
    \item A pipeline for generating a tetrahedral simulation-ready plant from real-world data.
    \item An FEM-based interactive plant dynamic model.
    \item An active branch manipulation motion planner that obeys deformation constraints.
    \item Open-source release of the developed software and datasets [wvu-irl.github.io/branch-modeling].
\end{enumerate}

These contributions enable gentle, active branch manipulation, making agricultural products easier to see and reach, thereby addressing critical limitations in harvesting, pruning, and pollination. 

The rest of the paper is outlined as follows: Section II discusses relevant works; Section III defines the problem statement; Section IV details the technical approach; Section V describes the experimental setup along with experimental results and discussion; Lastly, Section V concludes the work and outlines the potential future work of the study.

\section{Related Work}

\subsection{Branch Modeling}
A substantial body of prior work focuses on generating and reconstructing realistic branch geometry. Classical procedural models, such as L-systems~\cite{prusinkiewicz2012algorithmic}, space-colonization algorithm~\cite{runions2007modeling} and rule-based graphics pipelines~\cite{weber1995creation}, 
remain widely used for generating diverse tree topologies.

Recent research is shifting from hand-designed rules to estimating plant morphology directly from observations. Single-view learning methods can infer plausible tree structure from a single image~\cite{li2021learning}, while inverse/probabilistic formulations recover procedural parameters that best explain observed plant geometry from imagery~\cite{zhou2023deeptree} ~\cite{zhai2024cropcraft}. Complementarily, parametric plant models learned from real-world data provide compact representations that capture topology and shape variability across instances and species~\cite{Cheng2025Demeter}, enabling controllable reconstruction.

When dense 3D sensing is available, terrestrial laser scanning has enabled detailed structural recovery via quantitative structure models ~\cite{raumonen2013fast}, often by fitting cylinder-based representations that encode both geometry and branching hierarchy. Skeleton-first reconstruction pipelines further improve topological fidelity and detail; for example, AdTree~\cite{Du2019AdTree} and cherrypicker ~\cite{meyer2023cherrypicker} extract a tree skeleton to recover accurate 3D models from laser scans. Diffusion-based approach to generate tree skeleton from partially observed point clouds under occlusions~\cite{marks2025tree} is latest work that uses generative model to infer tree topology.

In parallel, forestry and plant biomechanics studies investigate mechanical response under environmental loading, including finite-element analyses that relate crown structure and material properties to sway and dynamic response~\cite{sellier2009crown}. While these works provide valuable physical insight, most geometry-centric reconstruction methods are not designed for interactive, contact-rich robot-plant manipulation, where loads and contact locations evolve online. This gap motivates integrating morphology reconstruction with physically grounded deformation modeling to support planning and control under active manipulation.

\subsection{Active Branch Manipulation}
Active branch manipulation aims to improve access to occluded targets (such as fruits, cut points, flowers) by intentionally interacting with and repositioning branches within cluttered canopies. Early pipelines combined 3D reconstruction with simplified dynamics models, representing branches as articulated links with spring-damper joints to predict canopy motion under applied forces \cite{yandun2020visual}. Related task-specific systems also model plant responses to interactions, such as pollination via controlled stem vibration, guided by physics-based discrete elastic rod simulations \cite{jeong2025vision}. These approaches are often computationally efficient, but their simplified dynamics can struggle to capture contact-rich, large-deformation behavior across complex, branched structures.

To enable simulation studies of contact-rich plant manipulation, some robotics simulators incorporate non-rigid plant models. For example, GazeboPlants extends Gazebo with Cosserat-rod-based plant dynamics to enable plant-robot interaction in simulation \cite{deng2024gazebo}. While such tools improve realism over rigid vegetation proxies, they still require calibration, and may not directly provide deformation-aware metrics tailored to safe manipulation planning.

On the control and planning side, prior work has explored interaction-aware strategies for operating in orchard-like clutter. Hybrid systems for pruning combine vision-based servoing with interaction control to reduce contact forces when establishing tool-branch contact \cite{you2022precision}. Reactive approaches, such as RICE, use tactile feedback to decide whether to maneuver around foliage or push through it, thereby improving robustness in deformable, occluded environments without requiring an explicit plant model \cite{Parayil2025rice}. While effective for reachability, these methods typically do not optimize motion with respect to deformation cost, and can therefore be conservative or apply unnecessary deformation depending on local canopy conditions. Online force-feedback based planning has been explored in  \cite{rijal2025force}, which minimizes the branch damage but requires multiple replaning attempts to bound force below safety threshold.  

Learning-based approaches have also been proposed to reason about occlusions and contact. Push Past Green learns to predict which actions will reveal free space beneath foliage, enabling sequences of actions that progressively expose occluded regions \cite{zhang2023push}. Other work learns forward deformation predictors and interaction policies from graph representations of spring-damper tree models \cite{kim2024towards}, branch-dynamics parameters via simulation-driven inference \cite{jacob2024learning}, or trains contact-aware policies for gentle branch interactions \cite{jacob2024gentle}. Reinforcement-learning-based planners have additionally been explored for occlusion-aware plant manipulation with sim2real transfer \cite{subedi2025find}. These methods can generalize to different branches, but may depend on simulator fidelity, reward shaping, or model abstractions that do not directly encode physically interpretable deformation measures to avoid branch damage.

In contrast to coarse elastic abstractions or purely reactive interaction, our work integrates a constitutive FEM branch model into planning and interactive manipulation. By using deformation measures derived from the FEM state (strain-energy-based costs) and updating estimates online during contact, we enable deformation-aware decision-making that remains physically grounded while supporting the responsiveness needed for real-world active branch manipulation.

\section{Problem Statement}

We represent a plant branch as a deformable body
$\mathcal{B} = \{ \mathbf{x}(s) \in \mathbb{R}^3 \mid s \in [0,L] \}$, where
$s$ denotes arc length and $\mathbf{x}(s)$ parameterizes the branch geometry.
In practice, branch shape, diameter, curvature, and branching structure vary
substantially across instances, leading to different deformation responses
under contact.

The deformation of the branch under an applied interaction force $\mathbf{f}$
is described by an unknown nonlinear mapping
$\mathcal{D} : (\mathcal{B}, \mathbf{f}) \rightarrow \Delta \mathcal{B}$, where
$\Delta \mathcal{B}$ denotes the change in branch configuration. This mapping
depends on spatially varying shape and material properties.

Given perception data $\mathcal{Z} = \{\text{RGB, depth, point cloud}\}$, the goal is to estimate the current branch model
$\hat{\mathcal{B}} = g(\mathcal{Z})$ and to design a manipulation strategy
$\pi : \hat{\mathcal{B}} \rightarrow \mathbf{\tau}$, where $\mathbf{\tau}$ denotes the
robot manipulation trajectory. The resulting
interaction should accomplish the task objective (such as repositioning an
occluded flower into field of view or into the reachable workspace) while satisfying
physical and safety constraints, such as bounds on contact force and allowable
deformation to avoid damaging the branch. 

We assume that the initial branch geometry can be fully captured at its rest configuration using a structure-from-motion pipeline \cite{schonberger2016structure}, yielding an initial point-cloud model. Since the focus of this work is on modeling and simulating branch deformation, we restrict attention to bare branches and do not consider leaf-induced occlusions. This assumption isolates the branch's deformation behavior and avoids the confounding effects of foliage on perception and interaction. Also, we assume a predefined grasp point $\mathcal{G}_t$ that remains fixed for each branch.

\section{Technical Approach}

This section describes the proposed approach for modeling and simulating branches from point cloud data, supporting waypoint generation while respecting deformation constraints. We also describe an iterative parameter estimation method to identify parameter values that yield realistic simulation behavior.

\subsection{Tetrahedral Plant Model}

We assume the input point cloud
$\mathcal{P} = \{ \mathbf{p}_j \in \mathbb{R}^3 \}_{j=1}^{N_c}$
represents a single plant and has been pre-segmented from the background.
To recover the underlying branch topology, we apply Laplacian-Based
Contraction (LBC)\cite{cao2010point}, which contracts the point cloud toward a one-dimensional
skeleton while preserving connectivity.
Given the original point cloud matrix $P \in \mathbb{R}^{N_c \times 3}$,
LBC iteratively solves the linear system
\begin{equation}
\label{eq:lbc}
\begin{bmatrix}
W_L L \\
W_H
\end{bmatrix}
P' =
\begin{bmatrix}
\mathbf{0} \\
W_H P
\end{bmatrix},
\end{equation}
where $L$ is the graph Laplacian of the point cloud, $W_L$ and $W_H$ are
diagonal weighting matrices controlling smoothness and positional fidelity,
and $P'$ denotes the contracted point positions. Repeated application of
(\ref{eq:lbc}) yields a skeletal point set
$\mathcal{S} = \{ \mathbf{s}_i \in \mathbb{R}^3 \}_{i=1}^{N_s}$
that approximates branch centerline and junctions.

After skeleton extraction, both the point cloud and skeleton are rigidly
aligned to a common upright coordinate frame such that the principal growth
direction coincides with the global $z$-axis. Let $\mathbf{R} \in SO(3)$ and
$\mathbf{t} \in \mathbb{R}^3$ denote the alignment transform; the aligned points
are given by $\tilde{\mathbf{p}}_j = \mathbf{R}\mathbf{p}_j + \mathbf{t}$ and
$\tilde{\mathbf{s}}_i = \mathbf{R}\mathbf{s}_i + \mathbf{t}$.
This normalization ensures consistent orientation and simplifies subsequent
geometric reasoning.

\begin{figure}[t]
    \centering
    \includegraphics[trim=0cm 1cm 5cm 3cm,clip,width=0.49\textwidth]{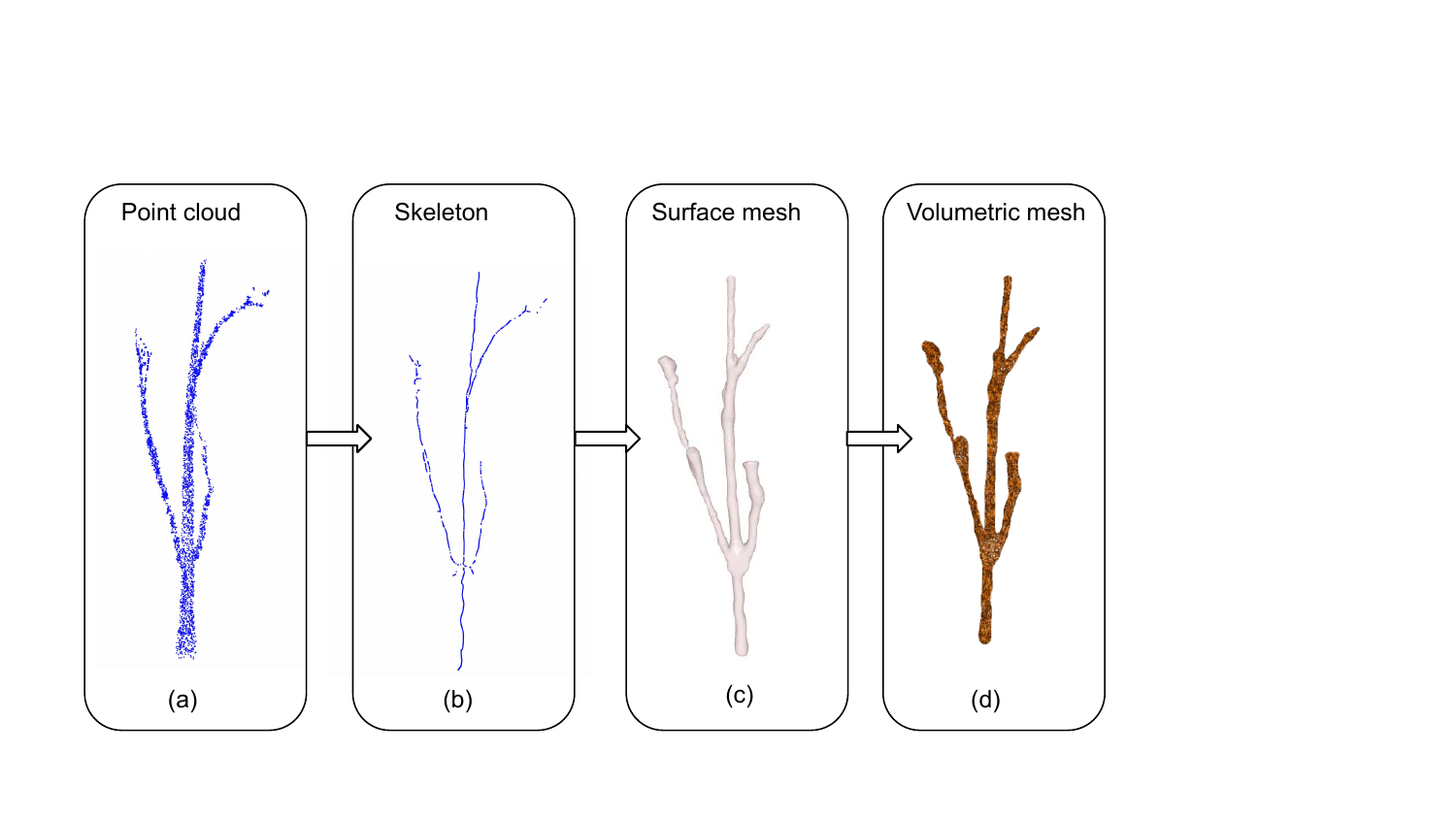}
    \caption{(a) Point cloud of the branch (b) Skeleton of point clouds generated by Laplacian contraction (c) Surface reconstructed from point cloud and skeleton (d) Volumetric tetrahedral mesh constructed from surface for simulation}
    \label{fig:overall_pipeline}
\end{figure}

Local branch radii are estimated from the aligned point cloud.
For each skeleton point $\tilde{\mathbf{s}}_i$, we compute its $k$-nearest
neighbors $\mathcal{N}_k(\tilde{\mathbf{s}}_i)$ and define the radius as
$r_i = \operatorname{median}_{\mathbf{p}_j \in \mathcal{N}_k(\tilde{\mathbf{s}}_i)}
\|\mathbf{p}_j - \tilde{\mathbf{s}}_i\|$.
To enforce physically plausible thickness, radii are clamped using the point
cloud scale $D$ as $r_i \in [\alpha D, \beta D]$, with $\alpha \ll 1$ and
$\beta < 1$.

Surface reconstruction is performed by fitting generalized cylinders along
the skeleton rather than using global surface reconstruction methods.
We found that methods such as ball-pivoting\cite{bernardini1999ball} can  produce fragmented, non-watertight
meshes, while Poisson reconstruction\cite{kazhdan2006poisson} can introduce bulbous artifacts due to
rapidly varying surface normals. Skeleton continuity is analyzed by detecting large inter-point gaps;
a new segment is initiated when
$\|\tilde{\mathbf{s}}_{i+1} - \tilde{\mathbf{s}}_i\| >
\gamma \cdot \operatorname{median}_k \|\tilde{\mathbf{s}}_{k+1} - \tilde{\mathbf{s}}_k\|$,
where $\gamma$ is a fixed gap multiplier.

For each continuous segment, a radius-adaptive generalized cylinder is
constructed by placing circular cross-sections orthogonal to the local tangent
$\mathbf{t}_i =
(\tilde{\mathbf{s}}_{i+1} - \tilde{\mathbf{s}}_i) /
\|\tilde{\mathbf{s}}_{i+1} - \tilde{\mathbf{s}}_i\|$.
Adjacent cross-sections are connected using triangular faces, yielding an
explicit branch-aligned surface mesh.

Finally, the reconstructed surface mesh is converted into a volumetric
representation using fTetWild\cite{hu2020ftetwild}, producing a high-quality tetrahedral mesh
suitable for physics-based simulation. This tetrahedral plant model preserves
branch topology and local thickness while enabling stable simulation suitable to visualize robot-plant interaction.

\subsection{Interactive Simulation}

After obtaining a simulation-ready tetrahedral plant model, we perform
interactive deformation simulation using the finite element method (FEM).
Plant structures are slender, highly deformable, and mechanically delicate,
exhibiting large bending and rotation under relatively small external forces.
Capturing such behavior requires a volumetric representation that can model
spatially distributed deformation and internal stress, which motivates the
use of FEM over other  model abstractions.

The branch dynamics is described by a second-order nonlinear system of
equations of motion,
\begin{equation}
\label{eq:fem_dynamics}
\mathbf{M}\ddot{\mathbf{u}} +
\mathbf{D}\dot{\mathbf{u}} +
\mathbf{f}_{\mathrm{int}}(\mathbf{u}) =
\mathbf{f}_{\mathrm{ext}}+ \mathbf{BC},
\end{equation}
where $\mathbf{u} \in \mathbb{R}^{3N}$ is the nodal displacement vector of the
tetrahedral mesh, $\mathbf{M}$ is the mass matrix, $\mathbf{D}$ is a damping
matrix, $\mathbf{f}_{\mathrm{int}}(\mathbf{u})$ denotes internal elastic forces, $\mathbf{f}_{\mathrm{ext}}$ represents externally applied forces such as
gravity, contact, or robot interaction and $BC$ represents boundary conditions that are imposed by
fixing nodes at the plant base or attachment locations, reflecting rooted or
supported growth.

Branch elasticity is modeled using the St.\ Venant-Kirchhoff (StVK)
constitutive model\cite{sifakis2012fem}, which captures large geometric deformations while
remaining computationally efficient. Let $\mathbf{F}$ denote the deformation
gradient and $\mathbf{\varepsilon} = \frac{1}{2}(\mathbf{F}^\top \mathbf{F} - \mathbf{I})$
the Green-Lagrange strain tensor. The corresponding strain energy density is
\begin{equation}
\label{eq:stvk_energy}
W(\mathbf{\varepsilon}) =
\frac{\lambda}{2}\big(\mathrm{tr}(\mathbf{\varepsilon})\big)^2
+ \mu\,\mathrm{tr}(\mathbf{\varepsilon}^2),
\end{equation}
where $\lambda$ and $\mu$ are the Lam\'e parameters. Internal elastic forces are
obtained by differentiating the total strain energy with respect to the nodal
displacements. This formulation allows large rotations and bending, which are
dominant deformation modes in branches and stems, while assuming moderate
strain levels consistent with the mechanical behavior of many branch tissues.

Time integration of (\ref{eq:fem_dynamics}) is performed using an implicit
Newmark scheme, which is well suited for stiff, bending-dominated systems.
Unlike explicit methods that require very small time steps for stability, the
implicit formulation enables stable simulation under large deformations and
external interaction forces. This allows the use of time steps appropriate for
interactive simulation, resulting in smooth and physically consistent plant
motion during robotic manipulation.

\subsection{Branch Parameters Estimation}

Accurate branch simulation requires realistic material parameters as these properties vary widely across plants (species, age, hydration) and
even along a single branch. To reduce manual tuning, we estimate the material parameters directly from real branch-robot interaction data, using
measured contact forces and observed plant motion. This is inspired by prior deformation capture and modeling pipeline\cite{wang2015deformation} that calibrates soft materials parameters from visual feedback and fit simulated models to observed deformations during interaction. 

\begin{figure}[t]
    \centering
    \includegraphics[trim=4cm 1cm 6cm 3cm,clip,width=0.48\textwidth]{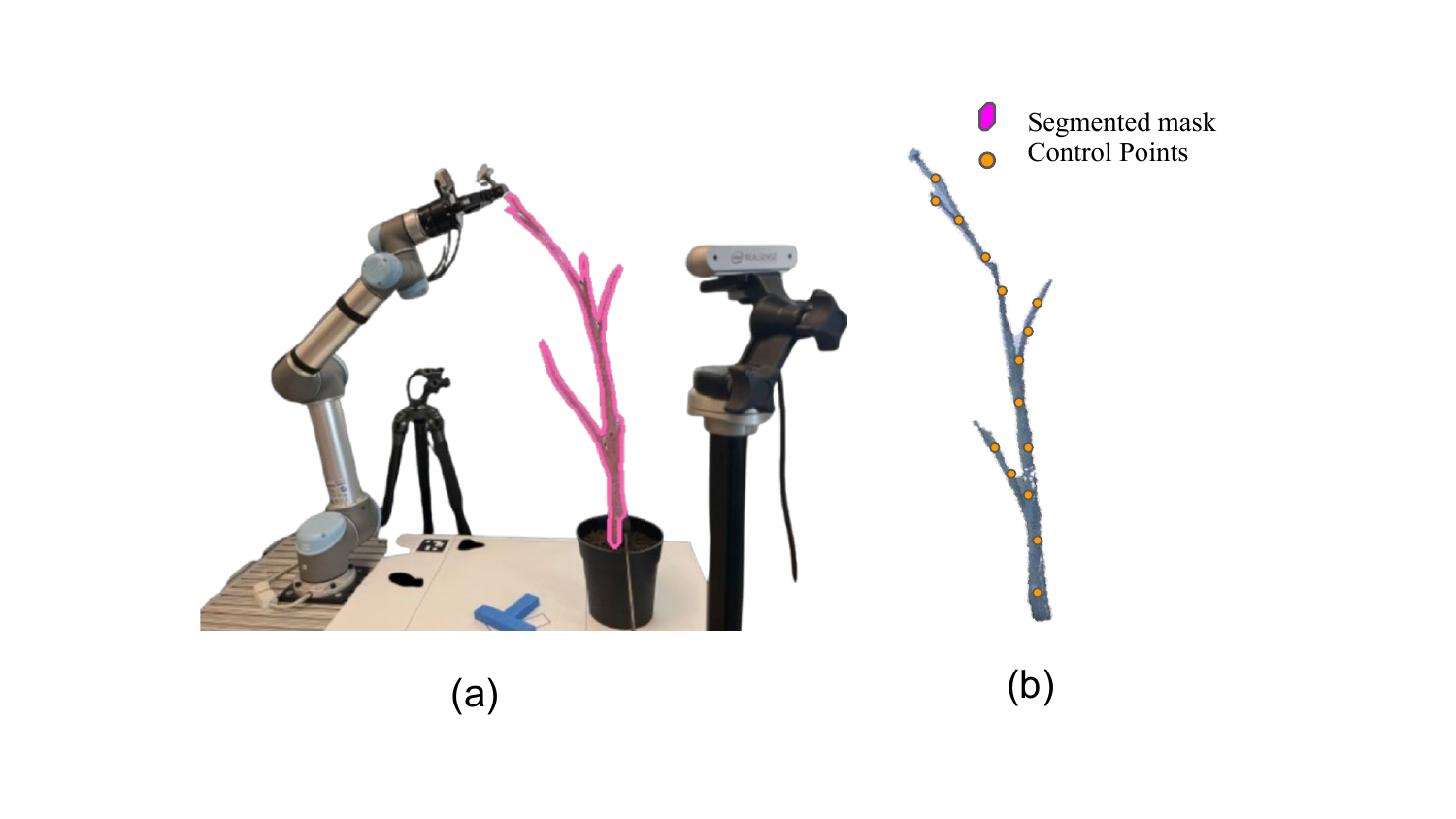}
    \caption{Branch parameter estimation setup where branch is manipulated by calibrated force and the corresponding deformed state is recorded (a) Segmentation mask of the branch (b) Point cloud obtained from segmentation mask and the depth value along with the control points shown as orange dots.}
    \label{fig:exp_mono}
\end{figure}

The robot applies controlled pulls at known
contact locations while a depth camera records the plant motion over time. With the help of segmentation mask and depth value point clouds of the deformation state is captured. $k$ controlled points $x_{t,k}(c)$ are sampled linearly across the length of the branch. These points has corresponding nodes $\hat{x}_{t,k}(c)$ in the simulation which are selecting by tracking these points from the original rest shape.  In parallel, the robot measures the interaction forces using a
force/torque sensor, which is mapped to a set of nodal contact forces
$\mathbf{f}^{t}_{i,\mathrm{ext}} \in \mathbb{R}^3$ at contact nodes
$i \in \mathcal{G}_t$ (with $\mathcal{G}_t$ denoting the active grasp set at time
$t$). Force values and the deformations are time stamped to match the applied force and corresponding deformations. 

To estimate branch material parameters, we use a spatio-temporal loss that compares observed and simulated branch motion over the entire sequence. Unlike a spatial loss at a single time step, this approach captures both deformation and its evolution over time by accumulating errors across all tracked points and time steps.

Now, the spatio-temporal loss $J(\theta)$ is defined as
\begin{equation}
J(\theta)
= \sum_{t,k} \left\| \mathbf{x}_{t,k} - \hat{\mathbf{x}}_{t,k} \right\|^{2},
\label{eq:loss}
\end{equation}
where $t$ denotes the frame index, 
$k$ denotes the node index and $\theta(E_{k},\nu)$ is the parameter that needs to be estimated. 

\begin{algorithm}[h]
\caption{Branch parameters estimation}
\label{alg:branch_param_est}
\begin{algorithmic}[1]
\Require Branch model $g(\mathcal{Z})$ , measurements $\{x_{t,k}\}$, control points $\{c_t\}$

\State Initialize simplex:
$\theta_1=(E_{k(1,0)},\nu_{0})$,
$\theta_2=(E_{k(2,0)},\nu_{0})$,
....$\theta_{k+1}=(E_{k(k+1,0)},\nu_0)$
\State $i \leftarrow 1$

\Loop
    \State Compute $J(\theta)$ using~(\ref{eq:loss}), for $k=1,2,3....k+1$
    \State Order $\theta_1,\theta_2,..\theta_{k+1}$ such that
    $J(\theta_1)\le J(\theta_2)....\le J(\theta_{k+1})$
    \State $\theta_c \leftarrow (\theta_1+\theta_2)/2$
    \State Reflection:
    $\theta_r \leftarrow \theta_c+\alpha_{r}(\theta_c-\theta_{k+1})$
    \State Evaluate $J(\theta_r)$ using~(\ref{eq:loss})
    \State Update simplex (reflection / expansion / contraction / shrink)
    \If{$J(\theta_1) < \delta$}
        \State \textbf{return} $\theta_1=(E_{k}^*,\nu^*)$
    \EndIf
    \State $i \leftarrow i+1$
\EndLoop

\end{algorithmic}
\end{algorithm}

As our simulation is non-differentiable with respect to $\theta$, we adopt a gradient-free optimization strategy.
 Specifically, we use the Nelder–Mead simplex method to iteratively evaluate candidate parameters. For each $\theta$, we run a forward simulation with the measured contact forces, compute $J(\theta)$, and update$\theta$ until convergence (Algorithm~\ref{alg:branch_param_est}). The resulting parameters improve physical realism by minimizing the deformation discrepancy between the real branch and the simulated model across the entire trajectory.

\subsection{Deformation-Aware Motion Planning}

To generate plant-deformation-aware branch manipulation motions toward a goal region, we plan in the robot workspace using a deformation-aware variant of RRT$^*$ , which we refer to as (D-RRT$^*$ ). In this formulation, each node corresponds to an end-effector pose $\mathbf{x}\in SE(3)$ in workspace, and candidate goal poses are sampled within the reachable region, modeled as a hemisphere whose radius is bounded by the estimated branch length. Branch dynamics are incorporated by evaluating the feasibility of each local connection under the branch model and by assigning costs that reflect the resulting deformation.

We define the incremental cost of adding a new node $\mathbf{x}_{\mathrm{new}}$ from a neighbor $\mathbf{x}_{\mathrm{nbr}}$ as
\begin{equation}
C =
\alpha_{\mathrm{def}}\,C_{\mathrm{def}}(\mathbf{x}_{\mathrm{nbr}},\mathbf{x}_{\mathrm{new}})
+\alpha_{\mathrm{dis}}\,C_{\mathrm{dis}}(\mathbf{x}_{\mathrm{nbr}},\mathbf{x}_{\mathrm{new}}),
\label{eq:rrtstar_edge_cost}
\end{equation}
where $C_{\mathrm{dis}}(\mathbf{x}_{\mathrm{nbr}},\mathbf{x}_{\mathrm{new}})$ measures the workspace distance between the neighbor and the new node. The deformation term
$C_{\mathrm{def}}(\mathbf{x}_{\mathrm{nbr}},\mathbf{x}_{\mathrm{new}})$ quantifies the branch deformation induced along the local motion connecting the two nodes and is computed from the St.\ Venant-Kirchhoff strain energy (\ref{eq:stvk_energy}) of the branch model. The weighting coefficients $\alpha_{def}$ and $\alpha_{dis}$ trade off motion efficiency against distance and effort requirement.

Deformation energy varies according to the configuration of the grasp node as it is constrained and thus determines the displacement of other nodes from the rest shape. The unconstrained nodes always takes the position that minimizes their energy\cite{borum2015free}. Hence, the energy of the branch configuration is determined by the configuration of grasp point. 

To predict deformation energy for previously unseen configurations, we first run forward simulations to compute energies over a set of sampled grasp-point configurations. We then fit a radial basis function (RBF) interpolation \cite{wright2003radial} to this dataset and use the resulting model to estimate deformation cost for new configurations during the D-RRT$^*$ planning stage.

\section{Experiments, Results, \& Discussion}
In this section, the experimental setup, results and discussion of the branch parameters estimation and deformation-aware planning are presented.

\subsection{Experimental Setup}

\begin{figure}[t]
    \centering
    \includegraphics[trim=4cm 2cm 6cm 1cm,clip,width=0.48\textwidth]{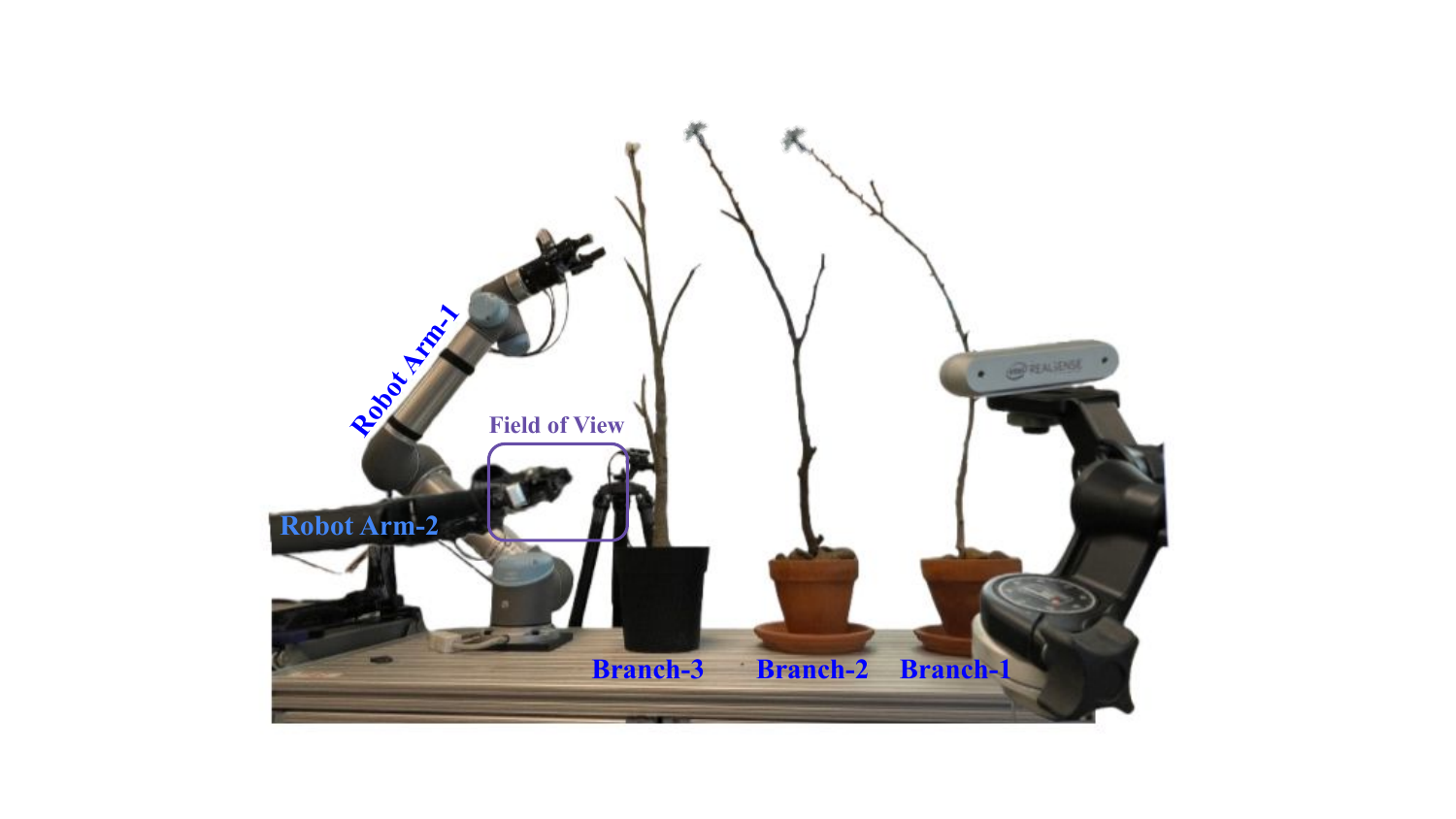}
    \caption{Experimental setup with three branches that are manipulated with arm-1 to move in the field of view of arm-2.}
    \label{fig:exp_setup}
\end{figure}

The experimental setup consists of two robotic arms and three branch types: young (Branch-1), mature (Branch-2), and artificial (Branch-3), each with an artificial flower at its tip (Fig.~\ref{fig:exp_setup}).

The first robotic manipulator is a UR5 equipped with a Robotiq FT300 force-torque sensor at the end-effector. It interacts with the selected branch at known contact locations to maneuver the tip-mounted flower into the reachable workspace of a second custom manipulator. Additionally, an Intel RealSense D455 camera provides RGB and depth images of the scene during all experimental trials.

\subsection{Branch Parameter Estimation}
\begin{figure}[h]
    \centering
    \includegraphics[trim=2cm 1cm 8cm 3cm,clip,width=0.48\textwidth]{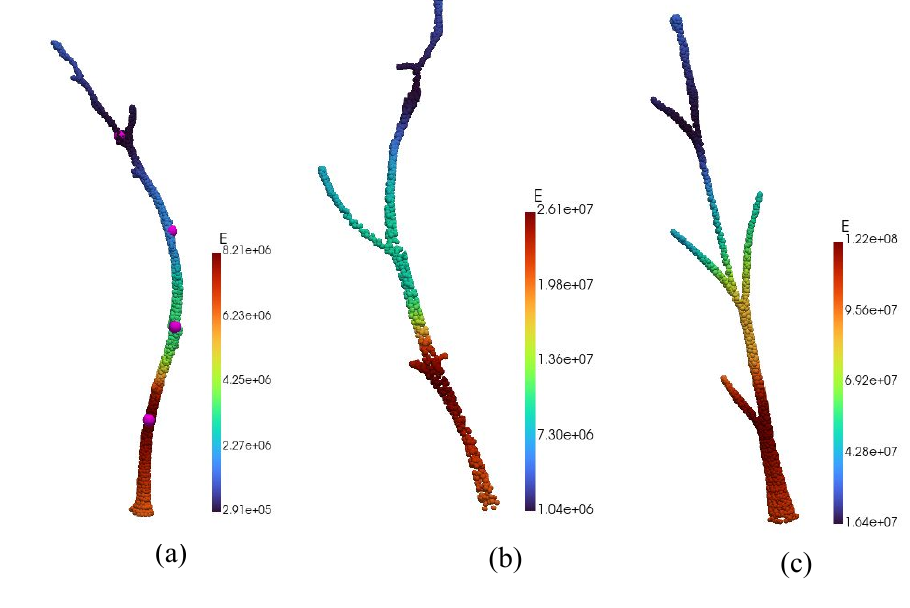}
    \caption{ Plot of $E$-values for different branches (a) young, (b) mature, (c) artificial  found after parameter estimation.}
    \label{fig:E_plot}
\end{figure}

In this experiment the branch material parameters are estimated for each branch type. Four ($K=4$) control points per branch are defined at fractions $\{1/5,\,2/5,\,3/5,\,4/5\}$ of the initial branch height, shown as magenta markers in Fig.~\ref{fig:E_plot}(a). The corresponding control points on the real plant are selected as the nearest visible surface point from the skeleton generated by the initial point cloud.

For each specimen, the UR5 manipulator executes a predefined trajectory while recording the end-effector waypoints and corresponding force measurements. Additionally, a partial point cloud of the branch is recorded by segmenting the scene camera's point cloud with a YOLO-26 model\cite{yolo26_ultralytics} (prompted with ``potted plant'') as shown in Fig.\ref{fig:exp_mono}(b). The measured forces are then applied to the simulated plant whose deformations are compared to the measured point cloud. Then Algortihm\ref{alg:branch_param_est} iteratively updates $\theta$ and terminates when the mean squared error between the simulated and observed control-point trajectories stabilizes below a preset threshold.

During simulation, it was empirically observed that varying Young's modulus ($E$) had a substantially larger effect on the branch deformation than Poisson's ratio ($\nu$). Therefore, each branch is modeled with a constant Poisson's ratio $\nu$, while allowing $E$ to vary across control points. Fig.~\ref{fig:E_plot} visualizes the resulting spatial distribution of Young's modulus for each of the branch specimen (young, mature, and artificial). The full-mesh $E$ field is obtained by smoothing the control-point values in Table~\ref{tab:estimated_material_params_1e7pa} using inverse-distance weighting interpolation\cite{shepard1968interpolation}.

\begin{table}[h]
\centering
\caption{Estimated parameters at four control points for each branch. Young's modulus values are reported in units of $10^{7}$ Pa.}
\label{tab:estimated_material_params_1e7pa}
\renewcommand{\arraystretch}{1.15}
\setlength{\tabcolsep}{4pt}
\begin{tabular}{lccccc}
\toprule
\textbf{Branch} & $\boldsymbol{E_1}$  & $\boldsymbol{E_2}$  & $\boldsymbol{E_3}$  & $\boldsymbol{E_4}$ & $\boldsymbol{\nu}$ \\
\midrule
Young & $0.821$ & $0.335$ & $0.182$ & $0.029$ & $0.382$ \\
Mature & $2.610$ & $0.913$ & $0.758$ & $0.104$ & $0.351$ \\
Artificial & $12.2$ & $8.60$ & $4.40$ & $1.60$ & $0.336$ \\
\bottomrule
\end{tabular}
\end{table}

\begin{figure}[h]
    \centering
    \includegraphics[trim=1cm 1cm 1cm 3cm,clip,width=0.49\textwidth]{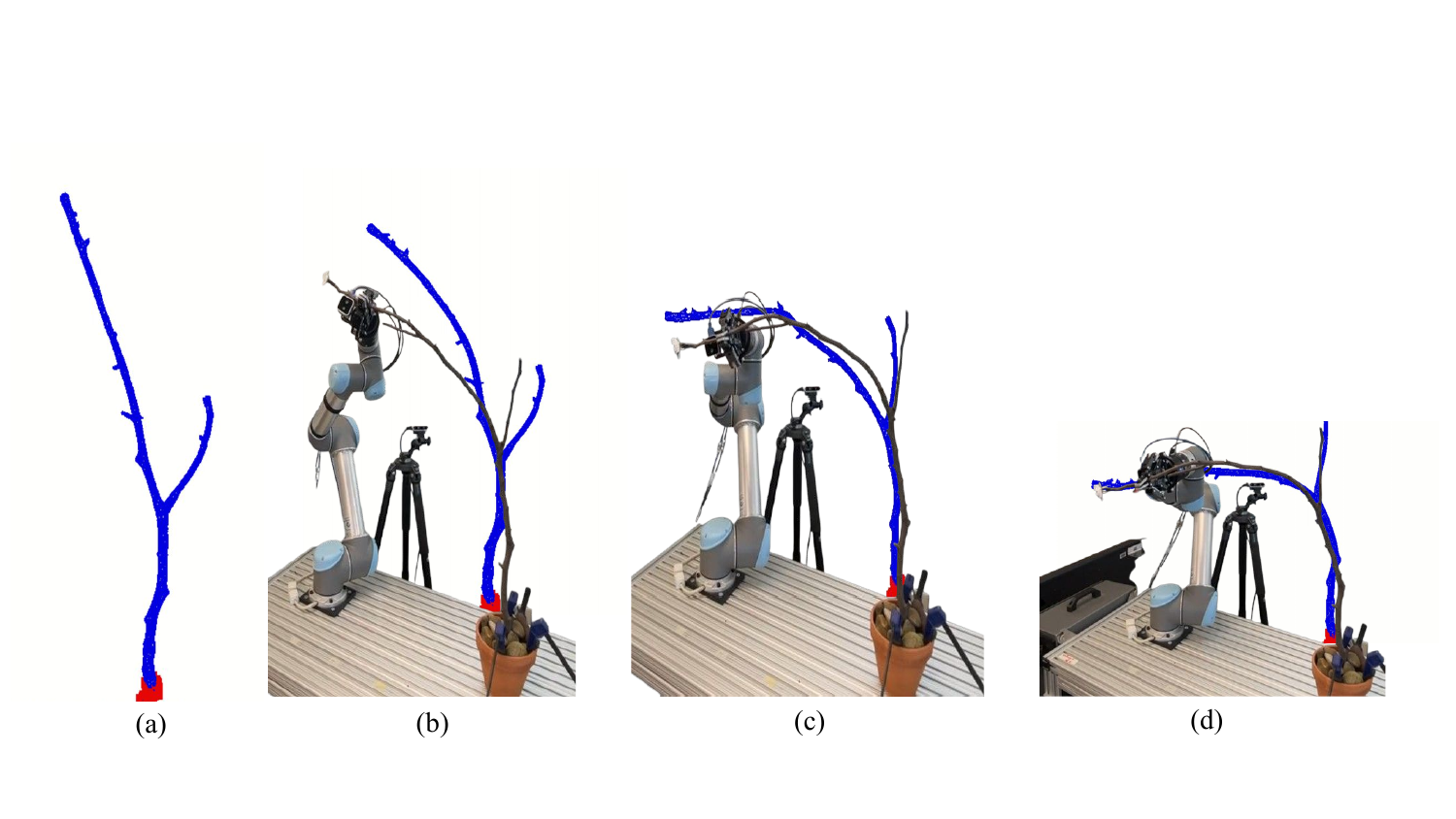}
    \caption{Deformation visualization of simulated (blue) and real mature branch (black) (a) rest shape of simulated branch, (b) error = 0.012$m^2$ (c) error = 0.0027$m^2$ (d) error = 0.0016$m^2$}
    \label{fig:sim_overlap}
\end{figure}

 These estimates induce a physically plausible stiffness gradient along the branch and reduce the deformation mismatch over the full interaction trajectory by allowing different regions of the plant to respond appropriately under the applied forces as shown in Fig\ref{fig:sim_overlap}. 
 
\subsection{Deformation Aware Planning}

\begin{figure}[h!]
    \centering
    
    \includegraphics[trim=1cm 1.5cm 6cm 2cm,clip,width=0.48\textwidth]{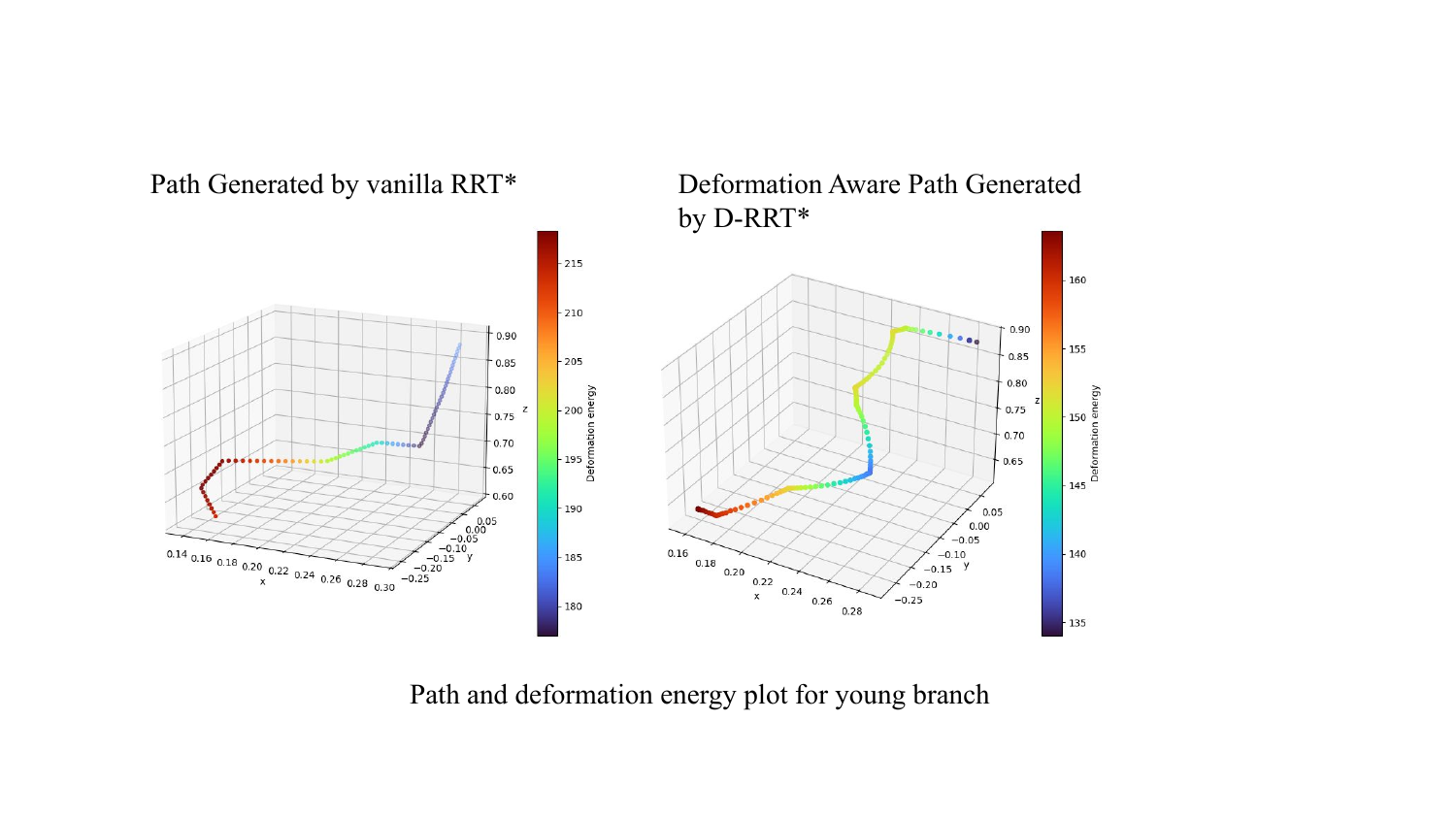}\par\vspace{10pt}
    \includegraphics[trim=1cm 1.5cm 5.5cm 2cm,clip,width=0.48\textwidth]{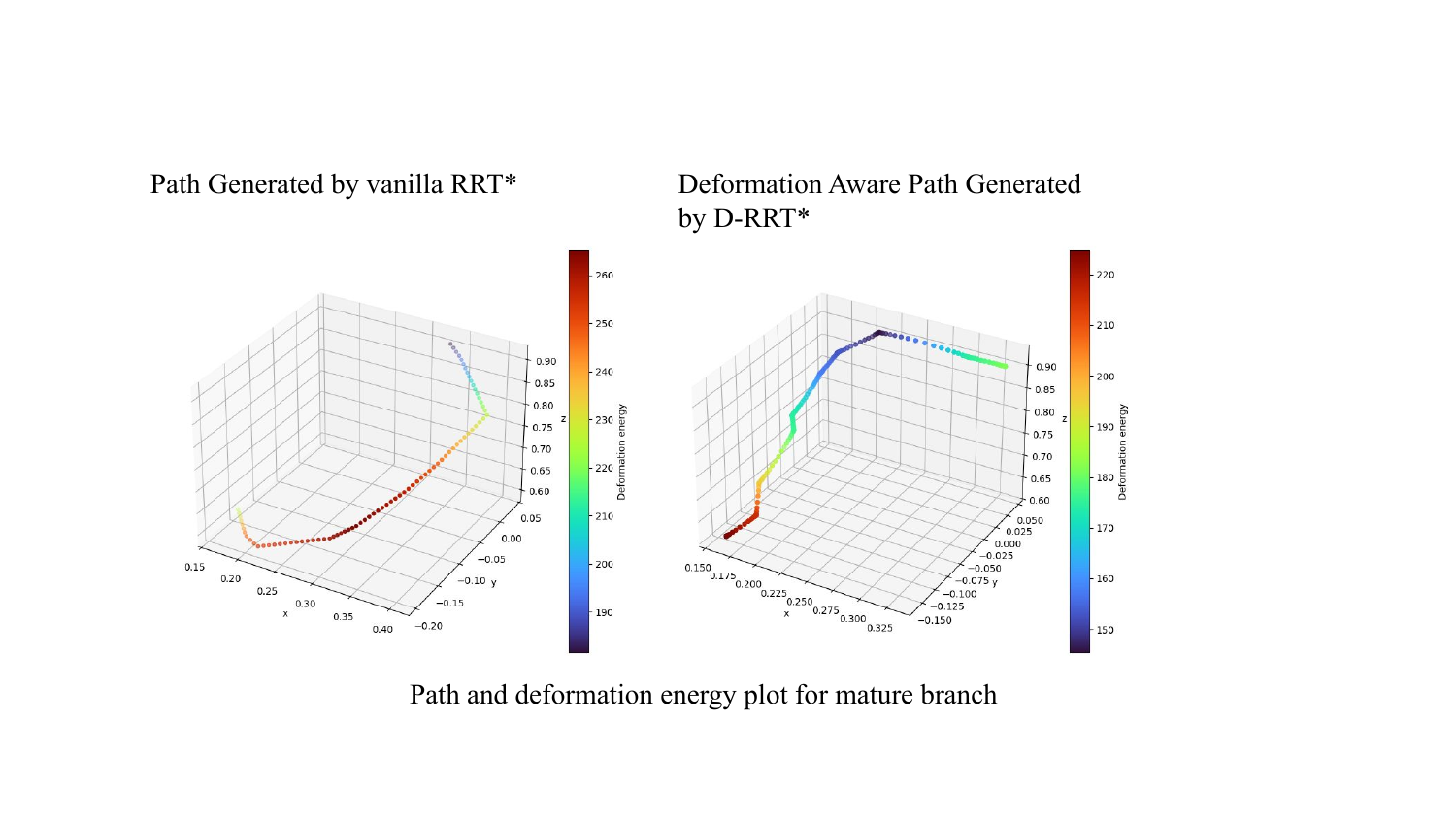}\par\vspace{10pt}
    \includegraphics[trim=1cm 1.5cm 5.5cm 2cm,clip,width=0.48\textwidth]{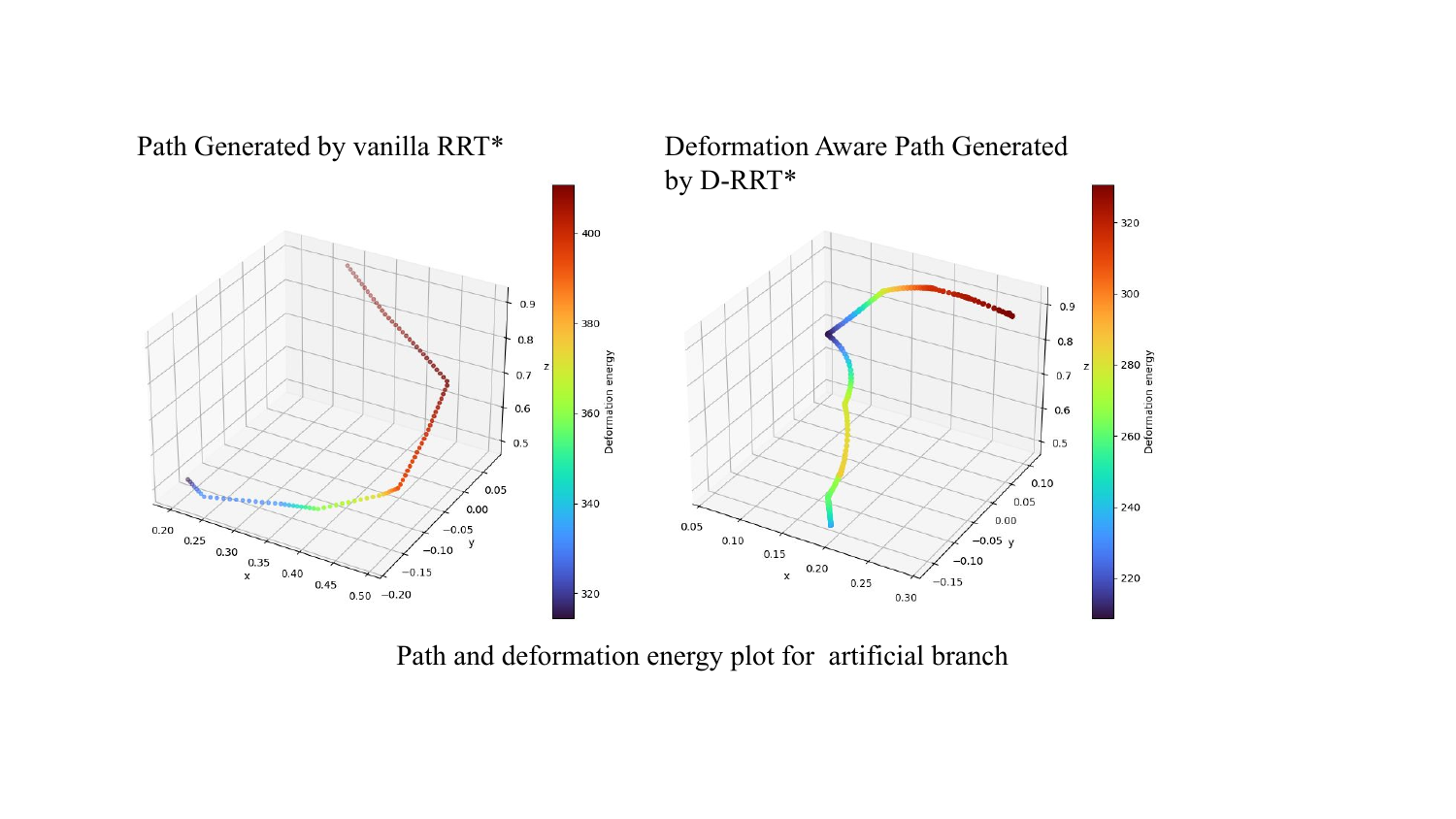}
    
    \caption{Visualization of deformation energy for paths generated by vanilla RRT$^*$ and D-RRT$^*$ across three branches. D-RRT$^*$ selects waypoints in lower-energy regions more consistently, while vanilla RRT$^*$ includes waypoints in higher-energy bands.}
    \label{fig:path_comparison_three}
\end{figure}

Fig.~\ref{fig:path_comparison_three} shows representative trajectories for each branch type along with the deformation energy values at each way-points. Across all the branches type, D-RRT$^*$ produces paths that are longer from the baseline RRT$^*$, typically taking deformation energy efficient routes to the same goal. This behavior reflects the planner's tendency to trade off path optimality for reduced deformation risk.

\begin{table}[h]
\centering
\caption{Comparison of baseline RRT$^*$ and deformation-aware RRT$^*$ (D-RRT$^*$) over 10 trials per branch. D-RRT$^*$ reduces deformation energy, at the cost of longer paths.}
\label{tab:rrt_vs_drrt_metrics}
\renewcommand{\arraystretch}{1.15}
\setlength{\tabcolsep}{3.5pt}
\resizebox{\columnwidth}{!}{%
\begin{tabular}{l cc c cc c}
\toprule
\textbf{Branch} &
\multicolumn{2}{c}{\textbf{Path Length (m)}} &
\textbf{$\uparrow$ (\%)}  &
\multicolumn{2}{c}{\textbf{Deformation Energy (N-mm)}} &
\textbf{$\downarrow$ (\%)}  \\
\cmidrule(lr){2-3}\cmidrule(lr){5-6}
& \textbf{RRT$^*$} & \textbf{D-RRT$^*$} & &
\textbf{RRT$^*$} & \textbf{D-RRT$^*$} & \\
\midrule
Young & $0.519\pm0.102$ & $0.552\pm0.134$ & 6.36  & $15821.66\pm1081.23$ & $10932.37\pm328.11$ & 30.90 \\
Mature & $0.538\pm0.106$ & $0.575\pm0.119$ & 6.88  & $19290.08\pm1324.42$ & $13044.17\pm734.26$ & 32.38 \\
Artificial & $0.647\pm0.168$ & $0.715\pm0.159$ & 10.51 & $30794.16\pm2183.79$ & $18409.94\pm721.61$ & 40.22 \\
\midrule
\textbf{Avg.} &
$-$ & $-$ & 8.10 &
$-$ & $-$ & 35.69 \\
\bottomrule
\end{tabular}%
}
\end{table}

To evaluate deformation-aware planning, the UR5 manipulator moves a branch from randomized initial configurations to the common reachable workspace of the custom manipulator, repeated for 10 trials per branch. The path generated by the proposed deformation-aware RRT$^*$ (D-RRT$^*$) is compared to the baseline vanilla version in terms of path length and strain energy induced in the branch as tabulated in Table \ref{tab:rrt_vs_drrt_metrics}. With an average path-length increase of 8.10$\%$, the deformation energy decreases by 35.69$\%$, suggesting that the planner can strategically steer the trajectory away from configurations that would otherwise incur high deformation costs.

\section{Conclusion and Future Work}
This paper presents an approach for active branch manipulation that leverages perception to build a branch model, simulate its deformation, and tune material parameters iteratively to reduce the discrepancy between observed and simulated motion. Experimental results shows that this method  can be generalized  to branches of varying geometries and material properties. Additionally, planning trajectories that account for branch deformation energy supports safer manipulation of delicate living materials.

This method aims to provide not only visual realism but also physically accurate behavior, since it models the full volumetric mesh of branches rather than only the surface mesh. The limitation of this method is increased computational cost compared to lightweight alternatives such as mass-spring models, which can be mitigated through model reduction and parallelization. Also, our current implementation does not account for occlusions caused by leaves, which can  be addressed by incorporating view point re-planning and part level segmentation of point clouds. 
 
 Future work in these directions includes learning branch breakage criteria for early risk-aware decisions, estimating parameters online during interaction and handling manipulation under reduced observability.

\section*{Acknowledgment}

We extend our gratitude to Dr. Nicole Waterland and her students for providing access to the greenhouse facility and plants.

\bibliographystyle{IEEEtran}

\bibliography{references}

\addtolength{\textheight}{-12cm}   

\end{document}